\documentclass{appolb}
\usepackage{epsfig}
\usepackage{epstopdf} 

\begin{document}
\title{Polish and English wordnets - statistical analysis of interconnected networks}
\author{Maksymilian Bujok, Piotr Fronczak, Agata Fronczak
\address{Faculty of Physics, Warsaw University of Technology,
Koszykowa 75, \\PL-00-662 Warsaw, Poland}}
\maketitle
\begin{abstract}

Wordnets are semantic networks containing nouns, verbs, adjectives, and
adverbs organized according to linguistic principles, by means of semantic
relations. In this work, we adopt a complex network perspective to perform a
comparative analysis of the English and Polish wordnets. We determine their similarities and show that the networks exhibit some of the typical
characteristics observed in other real-world networks. We analyse interlingual relations between both wordnets and deliberate over the problem of mapping the Polish lexicon onto the English one.

\end{abstract}

\section{Introduction}

A wordnet is a network of concepts which are connected according to their meaning by semantic relations. Its importance for natural language processing is already acknowledged through hundreds of projects and tools that relay on it \cite{wordnet}. The most important wordnet applications include word sense disambiguation \cite{Silva2012}, language teaching and translation \cite{Hu1998}, information retrieval \cite{Nie1996}, and document classification \cite{Peng2005}.

The first, and by far the best developed lexical database of this kind is the one created for the English language and developed at the University of Princeton - the so called Princeton WordNet (PWN) \cite{Miller1990}. It is commonly used as a reference for other lexical networks (several dozens of wordnets that follow PWN design are listed by the Global WordNet Associaton \cite{globalwordnet}).

\begin{figure}
\epsfig{file=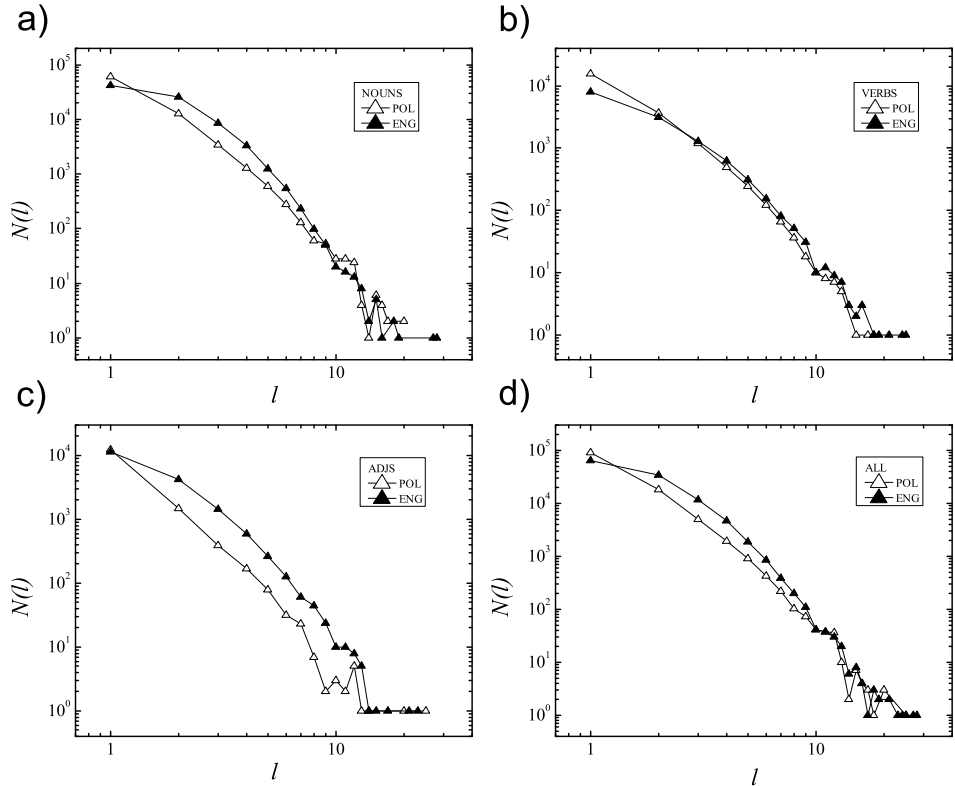,width=\columnwidth}

\caption{The number of synsets composed of $l$ lexems in both wordnets; a) nouns, b) verbs, c) adjectives, and d) all the grammatical categories.}
\label{fig1}
\end{figure}

Currently, huge efforts are being undertaken to map PWN onto other languages in aim to create a global wordnet grid as a one multilingual system \cite{Rudnicka2012}. For this reason, the understanding of the structure of particular wordnets, their differences and similarities, is not only interesting on its own right, as a challenging scientific problem. It is important because it provides an insight into the traps that researchers can fall into while semi-automatically matching objects from wordnets in unrelated languages.

In this paper we study several statistical characteristics of PWN and Polish wordnet (PolWN) \cite{Piasecki} using tools from the field of complex networks. We compare and discuss similarities and differences between both systems. Finally, we propose a method to assess the quality of mapping of both wordnets taking into account existing inter-lingual relations between them. We point out that due to differences in local topology of different wordnets the local network measures (e.g. node degree, clustering coefficient, motifs, etc.), which are frequently used in studying lexical networks \cite{Li2012,Steyvers2005,Ramanand2007,Borge2010}, can be misleading. 

\begin{figure}
\epsfig{file=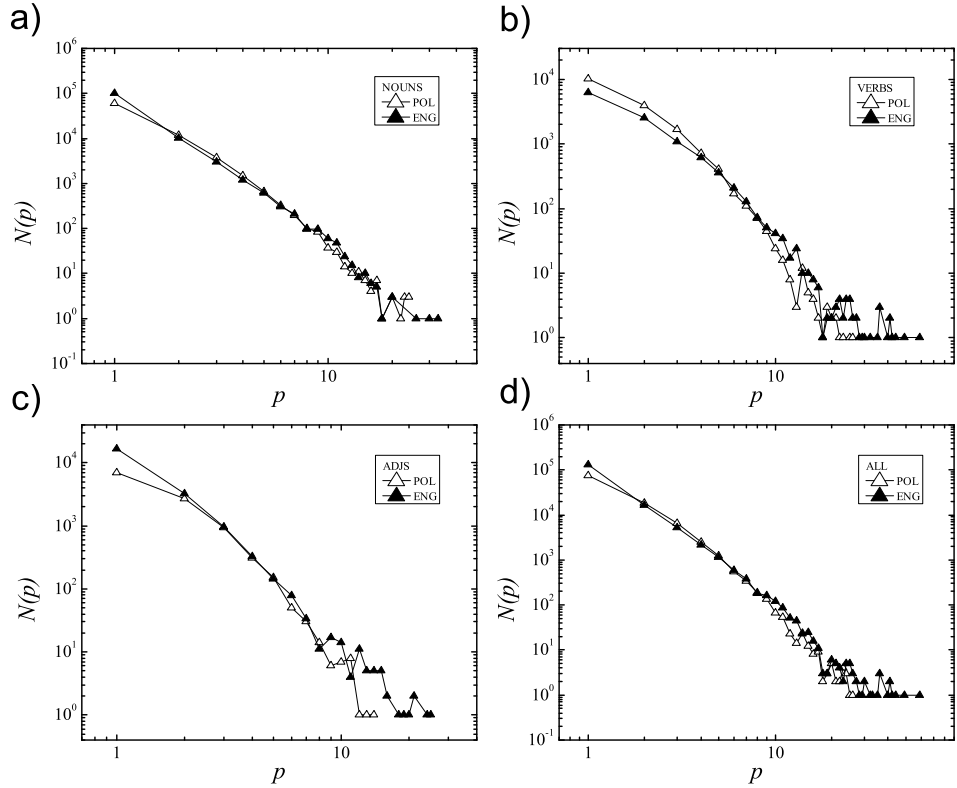,width=\columnwidth}

\caption{The polysemy distribution of the Polish and English lexems (the number of lexems that have the given number of senses); a) nouns, b) verbs, c) adjectives, and d) all the grammatical categories.}\label{fig2}
\end{figure} 

\section{Basic dataset statistics}

The basic building block of a wordnet is a \emph{synset}. Synsets are sets of synonyms that gather lexical items (\emph{lexems}) being interchangeable in a given context without changing the meaning. In the analyzed PWN and PolWN databases, there are $117659$ and $116319$ synsets, and $206978$ and $160098$ lexical items, respectively. It means that the average number of lexems in a synset is quite small (less than two). However, one should note, that in both wordnets, and for different grammatical forms (nouns, verbs, and adjectives), the synset size distributions are fat-tailed (see Fig. \ref{fig1}). Therefore, although more than half of all synsets are composed of only one lexem, the largest synsets contain $25$ (PolWN) and $28$ (PWN) synonymous lexems.  Despite apparent similarities between both lexicons, Fig. \ref{fig1}a and \ref{fig1}c may suggest that the synonymy in PolWN is more restrictive than in PWN \cite{Rudnicka2012} or that the Polish wordnet is still relatively less mature than PWN. 

Similarly, as a synset may contain many lexems (synonymy), a lexeme can participate in many different synsets (polysemy). Polysemy, the phenomenon when a single word has two or more meanings, is strictly related with the economy of the language, metaphoric thinking, imagery, and generalization \cite{Sigman2002}. On the other hand, the existence of polysemous words remains a central challenge to artificial intelligence in word sense disambiguation algorithms \cite{Ravin2000}. Fig. \ref{fig2} presents the polysemy distributions of the Polish and English lexems with respect to different grammatical categories. In all the cases, there are no significant differences between both wordnets. However in the range of extreme values, English polysemous lexems are more frequent. With respect to this remark, if one considers a wordnet as a network where lexems are connected when they belong to the same synset, then such extremely polysemous words act as hubs. It is well know, that highly connected nodes shorten distances within networks \cite{Fronczak2003}. Now, since distance in wordnets can be a useful proxy for semantic relatedness \cite{Budanitsky2006}, one can conclude that word sense disambiguation in the English lexicon, having more hubs, is more prone to errors. This hypothesis should be verified in a future.

\begin{figure}
\begin{center}
\epsfig{file=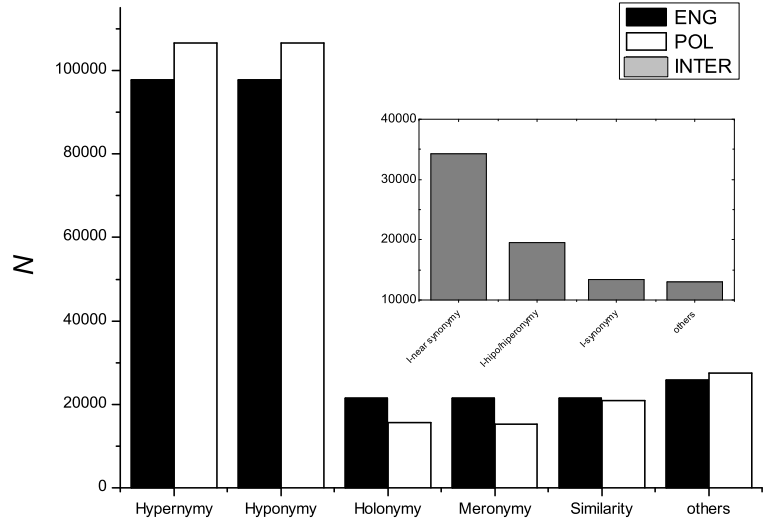,width=0.8\columnwidth}
\end{center}

\caption{Instances of semantic relations in the Polish and English wordnets. Inset shows the most important inter-lingual relations.}\label{fig3}
\end{figure}

\section{Wordnet as a complex network}
Synsets can be connected to other synsets via a number of semantic relations. The most important relationship is the hypernym/hyponym relation which bind a given concept to a more general/specific concept creating hierarchical connection, e.g. a \emph{dog} is a hyponym of \emph{animal} and \emph{animal} is a hyperonym of \emph{dog}. Another hierarchical relationship worth to mention is the meronym/holonym relation, which denotes part-whole/whole-part relationships, e.g. an \emph{engine} is a part of a \emph{car} (meronymy), and a \emph{car} contains an \emph{engine} (holonymy). There are several other relations (e.g. antonymy, casuality, etc.), but the semantic network is strongly dominated by the hypernymy/hyponymy links (see Fig. \ref{fig3}). That is why, in further analysis, we will take into account only this type of relationship.

\begin{figure}
\begin{center}
\epsfig{file=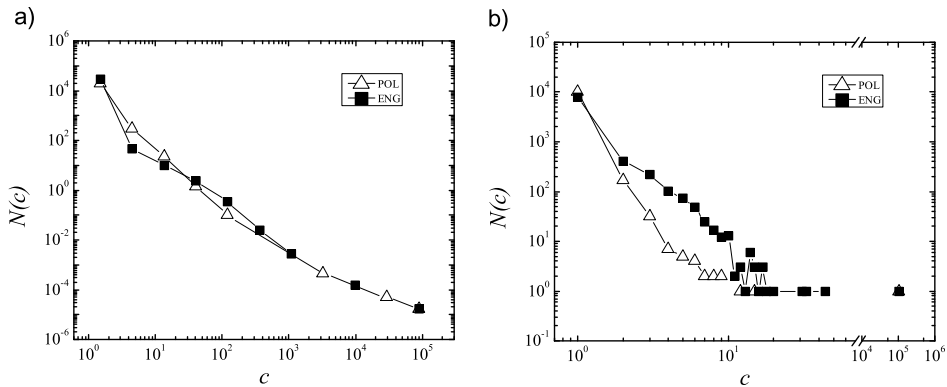,width=1\columnwidth}
\end{center}

\caption{Cluster size distributions for the Polish and English networks with a) hyperonymy links; b) all types of relations. Points in a) are averages over logarithmic bins along the horizontal axis.}\label{fig4}
\end{figure}

The synsets connected by hyperonymy links form an acyclic directed graph composed of a large number of isolated clusters. It is worth to mention that these clusters are not trees as was stated in the influential paper by Sigman and Cechi \cite{Sigman2002}. If one neglects directed character of the links, loops become quite frequent. The largest clusters in both wordnets (PolWN and PWN) consist of the nouns and contain respectively $59\%$ and $64\%$ of all synsets. The fat-tailed character of the cluster size distribution is shown in Fig. \ref{fig4}a. In this context both wordnets are very similar. To show the differences, one has to take into account all available relations. In that case, one can expect that the separated clusters merge together into the one giant cluster, i.e. there are no concepts isolated from the rest of the semantic network. Surprisingly, it is not true in both languages. Fig. \ref{fig4}b presents the cluster size distributions for the networks constructed as described above (with all relations). There exists the largest cluster (of size of $105661$ and $106236$ in PolWN and PWN respectively), but there are also many small-size clusters separated form the main core of the language. The most reliable explanation of this fact is that the construction of both wordnets is still not completed. If this explanation is correct, then, surprisingly, the younger Polish database outperforms the English one in that context.

\begin{figure}
\begin{center}
\epsfig{file=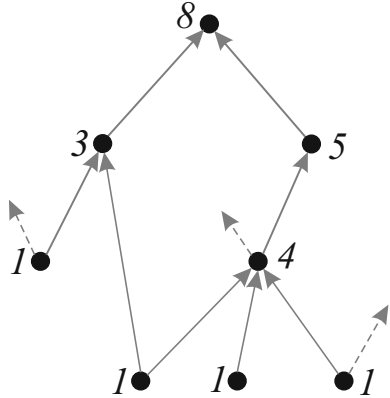,width=0.4\columnwidth}
\end{center}

\caption{Schematic illustration of the supremacy measure.
Solid arrows represent connections within the supremacy area of the
top vertex, whereas dashed arrows express connections pointing outside that area. Numbers situated in the vicinity of the nodes represent their supremacies.}\label{fig5}
\end{figure}

\begin{figure}
\begin{center}
\epsfig{file=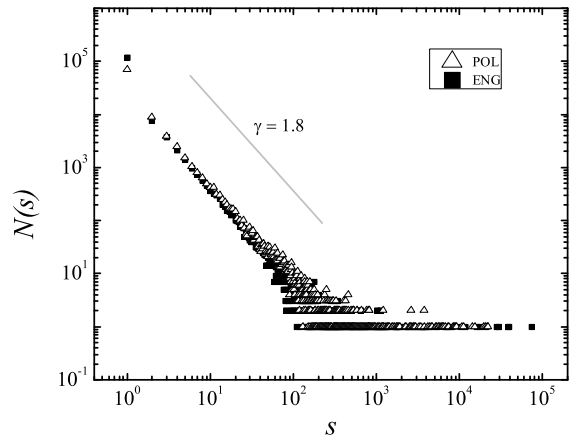,width=0.6\columnwidth}
\end{center}

\caption{Supremacy distributions in PWN and PolWN. Solid line represents the power law slope of both distributions.}\label{fig6}
\end{figure}

Importance of a synset can be estimated on several criteria, including the number of relations with other synsets, a position in the hyperonymic hierarchy of concepts \cite{Vossen1998}, or a sum of the frequencies of its component lexical units in the language corpus \cite{Buitelaar2002}. Since we have no access to the Polish corpus database, we decided to study synset importance with the help of the supremacy - a measure which encapsulates the first two criteria. Supremacy was proposed in \cite{Fronczak2004} as a parameter that can play an important role for description of a class of directed networks. It describes the number of nodes that are subordinated to a certain node. The parameter equals to the size of the basin connected to a certain node or to the size of its in-component and as such it has been used for the description of the Internet structure \cite{gam1} and
 for the scaling relations in food webs \cite{gam2}. An example of the network with the calculated supremacies of its nodes is presented in Fig. \ref{fig5}. With respect to this figure, please note, that the node with the supremacy equal to $5$ has less direct relations with other synsets than the node with the supremacy equal to $3$. Moreover, both nodes have the same position in the hierarchy. It demonstrates outperformance of the supremacy measure over standard criteria. Fig. \ref{fig6} presents the supremacy distribution for both wordnets. As in other complex networks, it is given by a power law. In both cases the characteristic exponent equals $1.8$. In the next section, we will take advantage of the similarity of both distributions when analyzing the quality of mapping PolWN onto PWN.

\begin{figure}
\begin{center}
\epsfig{file=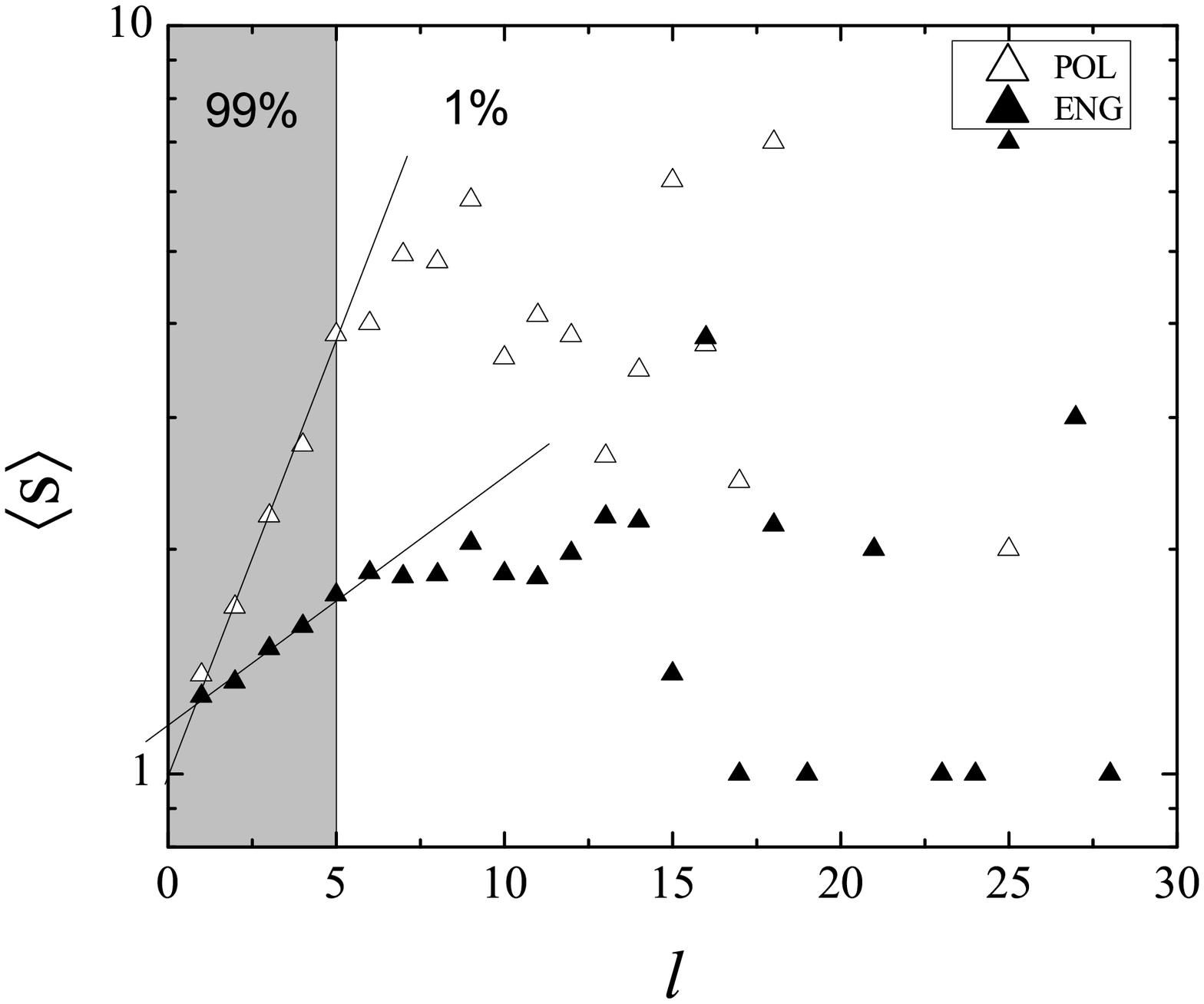,width=0.6\columnwidth}
\end{center}

\caption{The relation between the average supremacy of the synset and its size (i. e. the number of the lexems the synset is composed of). The straight lines are used to emphasize the linear character of the relation. Gray region contains more than 99\% of the synsets in both wordnets.}\label{fig7}
\end{figure}

In Fig. \ref{fig7} the relation between the average supremacy and the synset size (i.e. the number of the lexems the synset is composed of) is presented. Since the supremacy distribution is fat-tailed, instead of classical arithmetic averaging over the synsets with the same size, geometric averaging has been used. For the major parts of the lexicons (i.e. for more than 99\% of synsets in each wordnet)	there are clear linear relations between the logarithm of the supremacy and the synset size, i.e. $\langle s\rangle (l) \propto \exp(\alpha l)$. Interestingly, the scaling exponents $\alpha$ differ significantly between both wordnets ($\alpha_{POL}=0.26$ and $\alpha_{ENG}=0.08$). With respect to this observation, one can hypothesize that a new synonym, which enters into the synset when the language evolves, makes that concept much more important in Polish language than in English.

\begin{figure}
\begin{center}
\epsfig{file=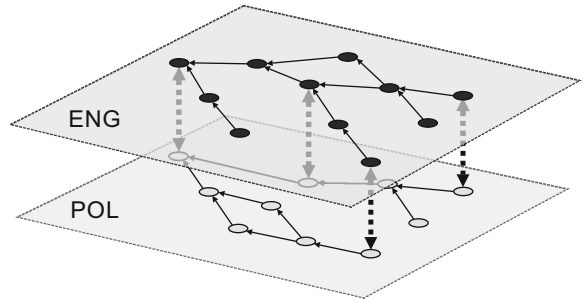,width=0.6\columnwidth}
\end{center}
\caption{The two-layered network composed of the Polish and English wordnets with the hiperonymy (thin arrows) and i-synonymy (thick dotted arrows) relations.}
\label{fig8}
\end{figure}

\section{The analysis of inter-lingual relations}
Interconnected network systems are of interest to researchers in numerous different disciplines \cite{Dorogovtsev2008,Buldyrev2010,Gao2011}. In linguistics the classical example of such a system is the EuroWordNet - the database that stores several European wordnets as autonomous language-specific structures that are interconnected via inter-lingual relations \cite{Vossen2004}. Since a wordnet-wordnet mapping holds great potential for crosslinguistic applications, the statistical analysis of the interconnected wordnets can shed a light on the commonalities and differences in the ways languages map concepts onto words. For example, it can help to detect language-specific lexical gaps, where a word in one language has no correspondence in another language \cite{Pease2010}.

Problems with the mapping of PolWN onto PWN arise from both the differences in the conceptual and lexico-grammatical structure of English and Polish languages, as well as from different methodological assumptions which underlie the construction of plWordNet where, on the contrary to PWN, lexems, instead of synsets, are the basic building elements \cite{Rudnicka2012}. 

Fig. \ref{fig8} presents the two-layered network composed of the Polish and English wordnets (directed acyclic graphs) connected by the inter-lingual synonymy (the so-called i-synonymy) relations. It is worth noting that, in general, the set of relations that connect both wordnets includes many different types of relations, such as inter-lingual synonymy and near-synonymy, inter-lingual hyponymy and hypernymy, and inter-lingual meronymy and holonymy, and some of them are much more frequent than i-synonymy (see Fig. \ref{fig3}). However, i-synonymy represents the most direct correspondence between the senses of the connected synsets. Therefore, for the sake of clarity we only take into account i-synonymy relations since they seem to be the most prone to matching errors.

\begin{figure}
\begin{center}
\epsfig{file=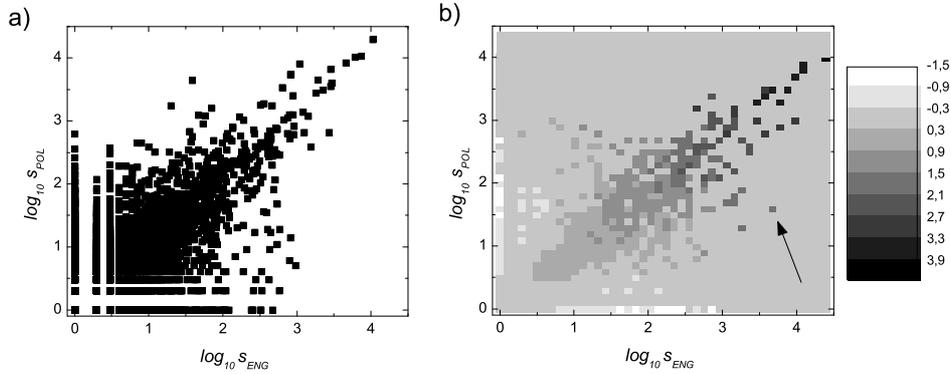,width=1.05\columnwidth,angle=0}
\end{center}
\caption{Supremacy profiles of the sysnets related by the inter-lingual synonymy. The colors represent the value of the ratio $R(s_{ENG},s_{POL})$, which detailed description can be found in the text.}
\label{fig9}
\end{figure}
The analyzed network contains $117659$ and $116319$ nodes, $89087$ and $106549$ directed links in the English and Polish layers respectively. Moreover, $L=13336$ i-synonymy links connect both layers. Now, it is reasonable to expect that the matched synsets share similar semantic importance in their own wordnets. Thus, when comparing their supremacies one should observe a clear correlation pattern. Each deviation from such a pattern can signal a potential mismatch of synsets or tell us about a language-specific gap in one of the two lexicons. Fig. \ref{fig9}a confirms that such a pattern exists and especially in the range of the large supremacies the correlation is clearly visible. On the other hand, in Fig. \ref{fig9}a, in the range of small supremacies a dispersion of points is high and one may get the impression that the topology of the i-synonymy connections is similar to that created in a completely random way. Such a thinking is however not correctly enough. To see this, let $L(s_{ENG},s_{POL})$ denotes the total number
of links connecting synsets with supremacies $s_{ENG}$ and $s_{POL}$.
This number can be compared to its typical value $L_0(s_{ENG},s_{POL})$ in the appropriate null-model network, i.e. the network with the randomly distributed i-synonymy links \cite{Maslov2003}. In such a network
\begin{equation}
L_0(s_{ENG},s_{POL})=p(s_{ENG})p(s_{POL})L,
\end{equation}
where $p(s)$ is the probability that a synset has a supremacy $s$ (the both probabilities, $p(s_{ENG})$ and $p(s_{POL})$, can be calculated, e.g., from Fig. \ref{fig6}). In Fig. \ref{fig9}b we illustrate the above comparison by plotting the ratio 
\begin{equation}
R(s_{ENG},s_{POL})= \log_{10}\frac{L(s_{ENG},s_{POL})}{L_0(s_{ENG},s_{POL})}.
\end{equation}

\begin{figure}
\begin{center}
\epsfig{file=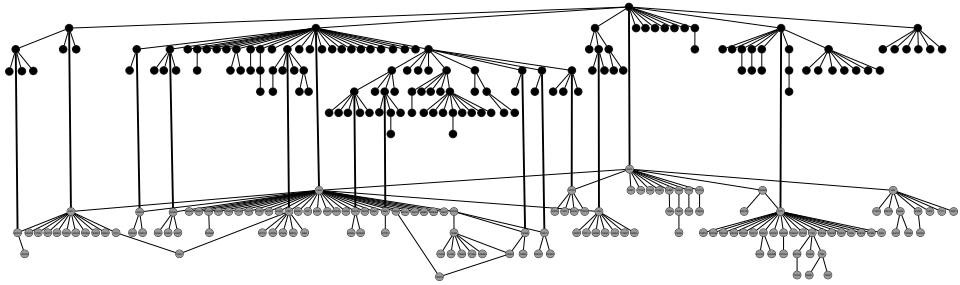,width=1.0\columnwidth,angle=0}
\end{center}

\caption{The in-components of the Polish synset \emph{publikacja} (gray nodes) and the English sysnet \emph{publication} (black nodes). Vertical lines represent inter-lingual synonymy relations.}\label{fig10}
\end{figure}

In our analysis, supremacies characterizing PWN and PolWN synsets are logarithmically
binned into five bins per decade. Regions, where $R(s_{ENG},s_{POL})$ is above (below) 0 correspond to enhanced (suppressed) connections between English and Polish synsets as compared to the randomized network. The presented procedure clearly demonstrates a linear pattern of correlations between the matched synsets. A better agreement for the larger values of the supremacy suggests that the matching has higher accuracy for the more general synsets. For example, Fig. \ref{fig10} presents two in-components of the Polish synset \{\emph{publikacja, wydawnictwo}\} and the English synset \{\emph{publication}\}. The supremacies of both synsets are $141$ and $140$ respectively. Thick vertical lines represent existing inter-lingual synonymy relations between the corresponding synsets. Although there is a visible similarity between both in-components, we would like to point out that they have different local topologies (which are probably due to different construction procedures of both wordnets). Thus, the two structures should not be compared on the level of node degrees, clustering coefficients, or other local measures as it was done in the previous studies (cf. \cite{Li2012,Steyvers2005,Ramanand2007,Borge2010}). The supremacy is an example of a measure which also takes into account more distant neighborhood of the node and as such it outperforms simple local measures. Finally, an example of the potentially incorrect mapping is marked by the arrow in Fig. \ref{fig9}. The Polish synset \{\emph{wiedza}\} has a supremacy $39$ while its matched English counterpart \{\emph{cognition, knowledge}\} - $4380$.

\section{Conclusions}

We have analyzed several statistical properties of the Polish and English wordnets with the help of the tools from the field of complex networks. We have proposed the supremacy as a simple but not yet trivial measure of the importance of synsets. Finally, we have used it for the analysis of the matching of both lexicons. It has been suggested that the presented methodology can be used to assess the matching quality or to find language-specific lexical gaps.

\section{Acknowledgements}

This work was supported by the Foundation for Polish Science (grant no. POMOST/2012-5/5).

\end{document}